\documentclass{article} %
\usepackage{iclr2026_conference,times}

\usepackage{amsmath,amsfonts,bm}

\def\eqref#1{equation~\ref{#1}}

\def\1{\bm{1}}

\DeclareMathAlphabet{\mathsfit}{\encodingdefault}{\sfdefault}{m}{sl}
\SetMathAlphabet{\mathsfit}{bold}{\encodingdefault}{\sfdefault}{bx}{n}

\usepackage{hyperref}
\usepackage{wrapfig}
\usepackage{url}
\usepackage{booktabs}
\usepackage{graphicx}
\usepackage{pgfplots}
\usepgfplotslibrary{fillbetween}

\title{Majority Voting for Code Generation}  %

\author{Tim Launer, Jonas Hübotter, Marco Bagatella, Ido Hakimi, Andreas Krause\\
ETH Zürich, Switzerland\\
}

\iclrfinalcopy %
\begin{document}

\maketitle

\begin{abstract}
We investigate Functional Majority Voting (FMV), a method based on functional consensus for code generation with Large Language Models, which identifies a representative solution from multiple generations using their runtime execution signatures on test inputs. We find that FMV is an effective test-time inference strategy, substantially boosting performance on LiveCodeBench without a large compute overhead. Furthermore, we extend the utility of functional consensus and apply it as an aggregation strategy for label-free Test-Time Reinforcement Learning. We demonstrate that this increases pass@1 on holdout tasks, but find no evidence of self-improvement beyond the base model's performance ceiling. \looseness=-1
\end{abstract}

\section{Introduction}

A central challenge in inference-time scaling of Large Language Models~(LLMs) for improved reasoning is discerning promising generations from hallucinations. While models can reliably generate plausible-looking chains of thought, their reasoning capabilities often remain brittle \citep{nezhurina2025alicewonderland, berglund2024reversalcurse}. In the absence of an external oracle verifier, recent approaches leverage strategies such as consensus sampling \citep{wang2023selfconsistency, gencluster2025, semanticvoting} or test-time reinforcement learning \citep{zuo2025ttrl} to improve performance at test time. However, applying these methods to code generation is a non-trivial challenge \citep{alphacode, li2025stesttimescaling}. Yet, code generation has a unique property: \textit{executability}. By executing generated programs against self-generated inputs, we can construct a ``functional consensus'' that serves as a proxy for ground truth \citep{mdr-exec, funcoder, menet2026coding}.

To this end, we investigate \textbf{Functional Majority Voting (FMV)}. While execution-based consensus has previously been employed for decoding \citep{mdr-exec} or as a component in recursive decomposition \citep{funcoder}, we isolate the soft functional voting mechanism itself as FMV to study its standalone scaling properties and its potential as a training signal. We demonstrate that this execution based signal serves a dual purpose:
\begin{itemize}
    \item \textbf{Standalone functional consensus for test-time inference:} We demonstrate that soft functional consensus significantly boosts performance over the baseline.
    \item \textbf{FMV-based Test-Time Reinforcement Learning:} We extend the utility of FMV beyond inference, using it as a pseudo-label generator for Test-Time Reinforcement Learning (TTRL) \citep{zuo2025ttrl} in two flavors. As a reward signal, we use the consensus execution vectors to verify rollouts (matching the consensus yields reward 1.0, otherwise 0.0).
\end{itemize}

\section{Related Work}
\label{sec:related-work}

\textbf{Test-Time Inference. } The foundation of our test-time inference strategy introduced in section \ref{sec:method} is the ``self-consistency'' decoding method introduced by \cite{wang2023selfconsistency}, who demonstrated that for complex reasoning tasks such as math, sampling a diverse set of reasoning paths and marginalizing over final answers significantly outperforms base performance. The intuition is that while incorrect reasoning paths are stochastic and diverse, correct reasoning paths tend to be consistent, leading to the same final answer.
Applying this consistency principle to code generation requires moving beyond text matching. While approaches like CodeT \citep{chen2022codet} rely on generating full input-output pairs, we simplify this by following \cite{mdr-exec}, requiring only generated inputs to establish functional consensus. Furthermore, \cite{funcoder} introduced functional consensus as a sub-component within \textit{FunCoder}, a recursive divide-and-conquer planning framework. While effective, their approach requires complex recursive problem decomposition. In contrast, we isolate the functional voting mechanism to study its scaling properties, demonstrating that execution consensus alone provides significant gains in both training and inference. \par
\begin{wrapfigure}{r}{0.33\textwidth}
    \vspace{-10pt} %
    \centering
    \includegraphics[width=\linewidth]{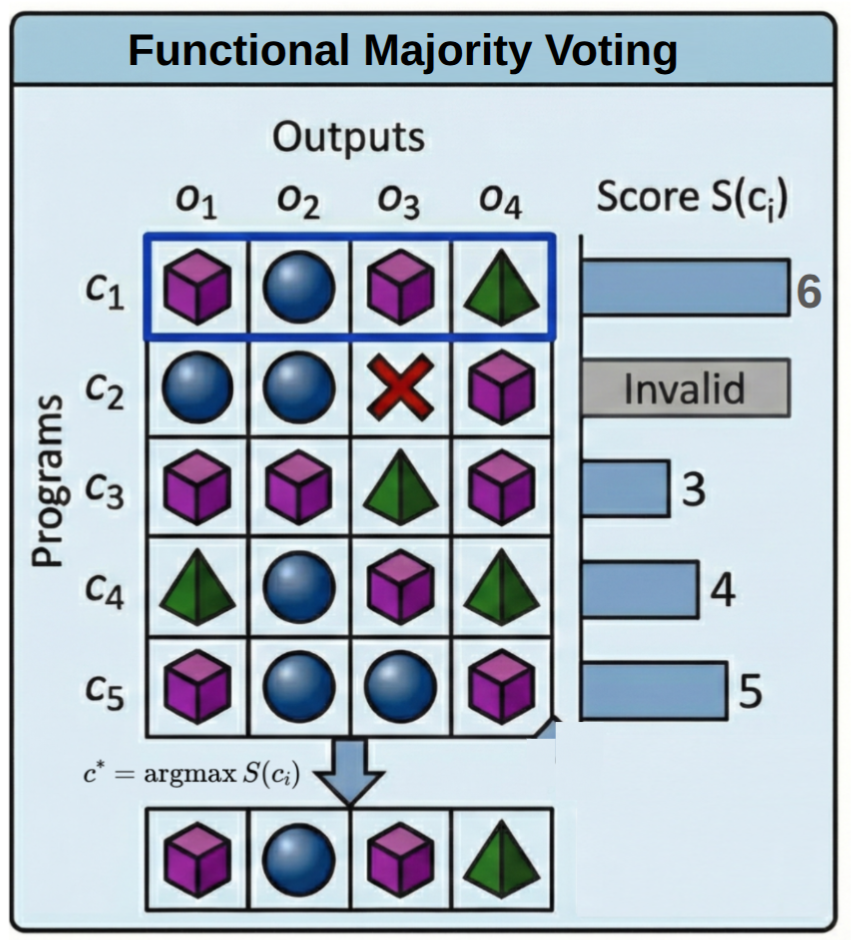}
    \caption{Candidate solution programs $c_1 ,\dots, c_5$ are sampled from a Large Language Model and evaluated against test inputs. Invalid candidates (e.g., $c_2$) are discarded. The consensus $c^*$ (here $c_1$) is selected by maximizing the Score $S(c_i)$.}
    \vspace{-20pt} %
    \label{fig:fmv-inution}
\end{wrapfigure}
Another execution-based clustering approach was recently employed by \cite{gencluster2025} in \textit{GenCluster} to achieve gold-medal performance on International Olympiad of Informatics (IOI) tasks. Similar to our method, they used execution based methods for finding promising candidates, though contrary to our approach they select final solutions using re-ranking (round-robin style) tournaments. \par
Finally, \cite{semanticvoting} introduce semantic voting, a method that relaxes hard matching consensus methods and instead uses soft matching based on semantic similarity. Specifically, each generated response is encoded into a vector using a sentence embedding model, and its voting score is computed as the average cosine similarity with all other responses. \par
\textbf{Test Time Reinforcement Learning. } Our investigation in Section~\ref{sec:training-results} is mainly inspired by \cite{zuo2025ttrl}, who introduced Test-Time Reinforcement Learning (TTRL), a method for training a pre-trained model at test-time using RL without ground-truth labels. To construct a reward signal, TTRL generates multiple candidate outputs from the model and identifies a consensus output via self-consistency, which is used as the ground-truth label for training. They demonstrate that for mathematical reasoning, a model trained with TTRL can solve problems far exceeding its initial baseline without any human supervision.

\section{Method: Functional Majority Voting}
\label{sec:method}

Let $\mathcal{M}$ be a model that generates $N$ candidate programs $\mathcal{C} = \{c_1, \dots, c_N\}$ given a prompt $P$. We assume access to a set of $K$ test inputs\footnote{If not available off the shelf, test case inputs can reliably be sampled from an LLM.} $\mathcal{X} = \{x_1, \dots, x_K\}$, but \textit{not} their ground-truth outputs. We execute every candidate $c_i$ on every input $x_k$ and collect the resulting output strings $o_{i,k}$ to obtain an execution vector $\mathbf{o}_i = [o_{i,1}, \dots, o_{i,K}]$. We define the set of valid candidates $\mathcal{C}_{val}$ as those that execute without runtime errors, invalid formats, or time outs on all inputs. The goal is to extract a consensus program $c^*\in \mathcal{C}_{val}$, representing the majority vote.

\textbf{FMV (The Functional Medoid\footnote{A ``medoid'' is the representative program with minimal disagreement to all other programs.}).} Paralleling \cite{mdr-exec} and \cite{funcoder}, we adopt a functional consensus scoring mechanism to select the most robust single consensus program $c^*$ for inference, and apply it directly to the candidate programs. Specifically, we identify the candidate that maximizes pairwise functional agreement with the ensemble of execution vectors. To this end, we define the \textit{FMV score} $S(c_i)$ for each $c_i \in \mathcal{C}_{val}$:
\begin{equation}
    S(c_i) = \sum_{j \neq i} \sum_{k=1}^{K} \mathbb{I}(o_{i,k} = o_{j,k})
    \label{eq:joint_fmv}
\end{equation}
The consensus program is then immediately given by $c^* = \mathrm{argmax}_{c_i} S(c_i)$ (see Figure \ref{fig:fmv-inution}).
Intuitively, $S(c_i)$ computes the average agreement across test cases between the output of the $i$-th candidate and those of other candidates. Standard consensus methods often require strict equivalence, meaning two programs must have identical output vectors to be grouped. This is brittle; if valid solutions differ on even a single edge case, strict voting fails to find a majority. To resolve this, this score uses a soft similarity metric that awards ``partial credit'' for matching on subsets of test cases.\looseness=-1

\textbf{Pointwise-FMV (Synthetic Targets).} For test-time training with TTRL \citep{zuo2025ttrl}, we can relax the requirement of selecting a single consensus candidate program. Therefore, as an alternative to using the execution vector to obtain the consensus program $c^*$ from above, we can instead construct a purely synthetic target execution vector $\mathbf{y}^*$: For each test case $x_k$, we select the mode of the output distribution: $y^*_k = \text{mode}(\{o_{1,k}, \dots, o_{N,k}\})$. Even if no single generated program solves all test cases, this allows us to construct a synthetic vector that represents a union of behaviors found in the ensemble, rather than being limited to the capabilities of the single best rollout.

\section{Results}
\label{sec:results}

Through our experiments, we use the \textbf{Qwen3} family \citep{qwen3} as our base models, evaluating both \texttt{Instruct-2507} (4B, 30B-A3B) and \texttt{Thinking-2507} variants. We evaluate on the code generation \textbf{LiveCodeBench-v6} (LCB-v6, \citet{livecodebench}). To isolate the voting mechanism from test-case generation quality, we use the oracle test inputs provided by the benchmark for executing, as well as their labels for evaluating our method. For each experiment, we report results with a rollout budget of $N=64$. Furthermore, for evaluating performance on unseen tasks after training, we split LCB-v6 into two randomized subsets of equal size, training only on one half. Finally, for evaluating Functional Majority Voting in TTRL \citep{zuo2025ttrl} in Section \ref{sec:training-results}, we restrict our analysis to \texttt{Qwen3-4B-Instruct-2507}. Hyperparameter settings are listed in Appendix \ref{sec:hyperparameters}.

\begin{figure}[t]
    \vspace{-2mm}
    \centering
    \begin{minipage}[c]{0.92\linewidth}
        \centering
        \includegraphics[width=\linewidth]{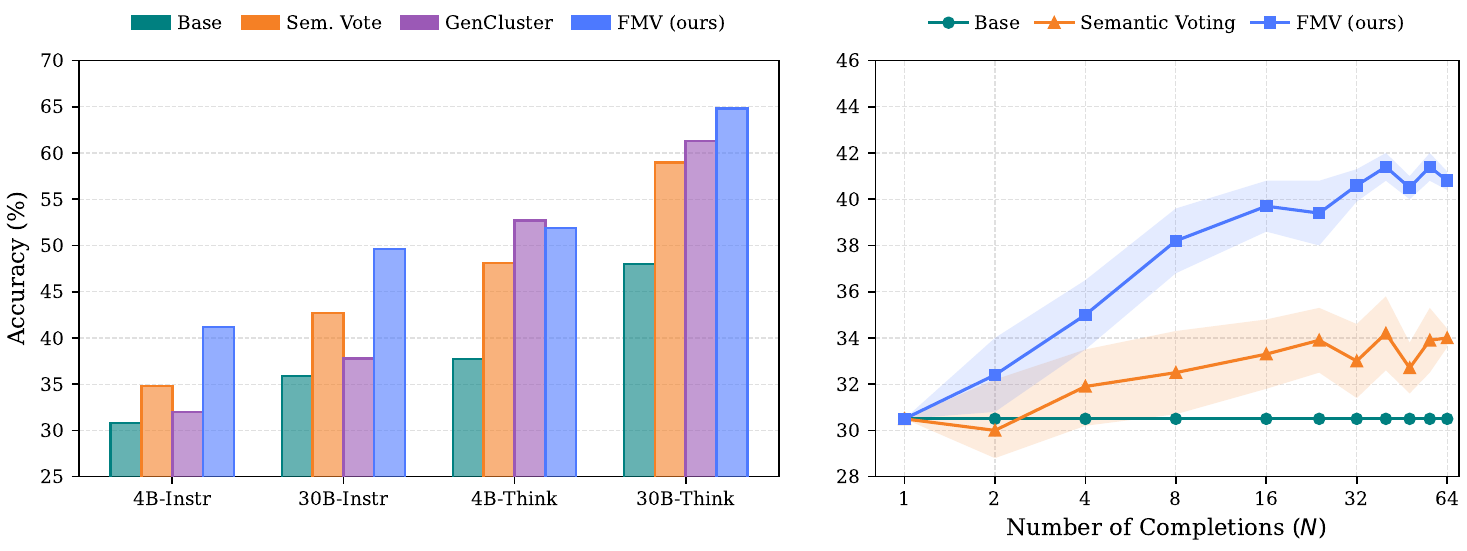}
        \label{fig:test-time-inference}
    \end{minipage}%
    \vspace{-6mm}
    \caption{FMV test-time inference evaluation. \textbf{Left:} Performance on LCBv6 ($N=64$) across model families. FMV consistently outperforms Semantic Voting and Base performance, and performs on par with GenCluster (Violet). \textbf{Right:} Scaling of voting methods for \texttt{Qwen3-4B-Instruct-2507} with rollout budget $N$ (log-scale). While Semantic Voting yields marginal gains over the Baseline, FMV scales efficiently, achieving over $40\%$ accuracy with as few as $N=32$ samples. Shaded regions show bootstrapped standard deviation of accuracy.}
    \vspace{-2mm}
    \label{fig:side_by_side}
\end{figure}
\subsection{Test-Time Scaling}
\label{sec:multimodel}

In this section, we evaluate the functional consensus mechanism \citep{funcoder} as a standalone test time scaling strategy: Functional Majority Voting (FMV). We compare the performance of FMV against the base model's pass rate, as well as against recent baselines: Semantic Voting \citep{semanticvoting} and GenCluster \citep{gencluster2025}. Figure \ref{fig:side_by_side}  (left) presents the results on LCB-v6 \citep{livecodebench} with a rollout budget of $N=64$. We observe that both Semantic Voting and GenCluster yield an improvement over the expected performance of a single sample ({mean@$64$}), increasing accuracy from $37.7\%$ to $48.1\%$ and $52.7\%$ respectively for \texttt{Qwen3-4B-Thinking-2507}. In comparison FMV (ours) consistently outperforms Semantic Voting and performs on par or better than GenCluster. Furthermore, we analyze the scaling properties of the consensus signal in Figure \ref{fig:side_by_side} (right), showing that FMV exhibits efficient and consistent improvement with the rollout budget $N$. %

Notably, FMV achieves these gains compute efficiently. Assuming an average of $K$ functional clusters per prompt, GenCluster relies on an additional $O(K)$ pairwise LLM-as-a-judge calls to rank clusters. In contrast, FMV relies purely on CPU-based execution, eliminating the latency and cost of auxiliary model calls, thus bringing significant efficiency gains.

\subsection{Self-Improvement via Test-Time Reinforcement Learning}
\label{sec:training-results}
\begin{table}[t]
    \centering
    \caption{TTRL performance on randomized training and holdout subsets of LCB-v6, showing improved performance on both. Errors shown correspond to bootstrapped standard errors of the mean. }
    \label{tab:fmv-ttrl-holdout}
    \begin{center}
    \begin{tabular}{lcc}
        \toprule
        \textbf{Method} & \textbf{Train (mean@64)}  & \textbf{Hold-Out (mean@64)} \\
        \midrule
        Base (Qwen3-4B-Instruct) & $30.8 \pm 3.2$ &  $31.6 \pm 3.1$ \\
        + FMV TTRL & $36.9 \pm 3.8$ & $34.3 \pm 3.2$ \\
        + Pointwise-FMV TTRL &  $36.6 \pm 3.8$ &  $34.5 \pm 3.2$ \\
        \bottomrule
    \end{tabular}
    \end{center}
\end{table}
Having validated FMV as a robust strategy for inference-time scaling, we assess its utility as a pseudo-label generator for Test-Time Reinforcement Learning (TTRL) \citep{zuo2025ttrl}. As discussed in Section \ref{sec:method}, both FMV and Pointwise-FMV can be used to construct the consensus reward signal for TTRL. In order to study the self-improvement behavior of FMV in TTRL, we track not only the single sample pass rate ({mean@$64$}), but also the maximum potential quality of the ensemble ({best@$64$}), which measures the percentage of questions with at least one fully correct response among $64$ attempts.\looseness=-1

\textbf{Generalization to Unseen Tasks.} We find that TTRL with FMV targets shows promise in generalizing to unseen tasks. Indeed, Table~\ref{tab:fmv-ttrl-holdout} shows that TTRL improves the base model's zero-shot performance from $31.6\%$ to $34.5\%$ on unseen problems, demonstrating that TTRL+FMV can improve zero-shot coding ability.

\begin{table}[t]
    \centering
    \caption{TTRL results on full LCBv6 for different FMV targets and training rollout budgets $N$. Synthetic FMV targets improve average performance over the base model but fail to raise the model's ceiling ({best@64}) or FMV accuracy, suggesting that TTRL is mainly amortizing inference-time scaling to zero-shot performance. Errors shown correspond to bootstrapped standard errors of the mean. }
    \label{tab:self-training}
    \begin{center}
    \begin{tabular}{lccc}
        \toprule
        \textbf{Training Method} & \textbf{mean@64} (\%) & \textbf{FMV} (\%) & \textbf{best@64} (\%) \\
        \midrule
        Base (Qwen3-4B-Instruct) & $30.8 \pm 3.2$ & $\mathbf{41.2} \pm 4.3$ & $\mathbf{48.9} \pm 1.0$ \\
        \midrule
        \multicolumn{4}{l}{\textit{Test-Time Reinforcement Learning}} \\
        + Joint ($N=32$) & $34.4 \pm 3.8$ & $38.9 \pm 4.1$ & $44.2 \pm 4.2$ \\
        + Pointwise ($N=32$) & $36.6 \pm 3.8$ & $40.5 \pm 4.3$ & $46.6 \pm 4.2$ \\
        + Joint ($N=128$) & $\mathbf{36.9} \pm 3.8$ & $40.5 \pm 4.2$ & $45.0 \pm 4.2$ \\
        + Pointwise ($N=128$) & $36.8 \pm 4.2$ & $38.9 \pm 3.9$ & $46.6 \pm 4.2$ \\
        \bottomrule
    \end{tabular}
    \end{center}
\end{table}

\textbf{Amortization of FMV gains.} Table~\ref{tab:self-training} shows that training on the full benchmark set with FMV targets successfully raises the zero-shot accuracy ({mean@64}) above the base model (e.g., $30.8\% \rightarrow 36.9\%$). However, the model fails to raise its own ceiling: the {best@64} score degrades consistently (e.g., $48.9\% \rightarrow 45.0\%$). Furthermore, applying FMV  on top of the trained model shows reduced relative performance gain after training, suggesting no evidence for recursive self-improvement. Instead, the model seems to be amortizing the gains of FMV as a test-time scaling method. This stands in contrast to gains reported in math domains \citep{zuo2025ttrl}. We attribute this limitation to a breakdown of the ``Lucky Hit'' phenomenon mentioned in \cite{zuo2025ttrl}: our models exhibit substantive agreement even on incorrect solutions, leading to a high rate of false positive rewards that reinforce errors rather than pruning them.

\section{Conclusion}

We study Functional Majority Voting, a consensus method that identifies representative solutions using runtime execution outputs. Our experiments validate that FMV serves as an effective standalone test-time inference strategy, substantially boosting performance on LiveCodeBench. Furthermore, we extend the utility of functional consensus as a labeling method in FMV-TTRL. We show that the method demonstrates increased zero-shot performance on both training and holdout tasks, but find that our experiments show no evidence for self-improvement beyond the base model's performance ceiling. \looseness=-1

\bibliography{iclr2026_conference}
\bibliographystyle{iclr2026_conference}

\appendix
\section{Hyperparameters}
\label{sec:hyperparameters}

\textbf{Training: } We perform training runs across two settings of the rollout budget per prompt, $N=32$ and $N=128$, in order to evaluate the scaling behavior of the method. No hyperparameter sweeps were performed, and we kept a fixed hyperparameter configuration with a temperature of $\tau=1.0$ for training, learning rate of $10^{-6}$ and constant warm-up, a training batch size of 8, gradient clipping of 1.0,  clip ratio of $[L,H]=[0.2, 0.32]$ and no KL loss term. Models are trained until performance on the validation set saturates.\par
\textbf{Inference: } During inference evaluation we use similar settings as \cite{wang2023selfconsistency}, using a temperature of $\tau=0.6$, $top_p=0.95$ and $\text{max\_new\_tokens}=8192$, $\text{max\_new\_tokens}=16384$ for \texttt{Instruct} and \texttt{Thinking} models respectively.

\end{document}